\documentclass[11pt,a4paper]{article}
\usepackage[hyperref]{acl2021}
\usepackage{times}
\usepackage{latexsym}

\usepackage[hyperref]{acl2021}
\usepackage{amsmath,amsthm,amssymb}
\usepackage{calrsfs}
\usepackage{times}
\usepackage{latexsym}
\usepackage{adjustbox}
\usepackage{hyperref}
\usepackage{float}

\usepackage{times}
\usepackage{xspace}
\usepackage{latexsym}

\usepackage{graphicx}
\usepackage{caption}
\usepackage{subcaption}
\usepackage{times}
\usepackage{url}
\usepackage{latexsym}
\usepackage{amsmath}
\usepackage{lipsum}
\usepackage{multirow}
\usepackage{booktabs}
\usepackage{enumerate}
\usepackage{dblfloatfix}
\usepackage{cleveref}
\usepackage{microtype}
\usepackage{amsmath}
\usepackage{colortbl}

\DeclareMathOperator*{\maxB}{max}

\usepackage{tikz}
\usepackage{amsmath}
\usepackage{fixltx2e}
\usetikzlibrary{shapes.geometric, arrows}

\tikzstyle{input} = [rectangle, minimum width=0cm, minimum height=0.6cm,text centered, draw=white]
\tikzstyle{dummy} = [minimum width=0cm, minimum height=0cm, draw=white]
\tikzstyle{block1} = [rectangle, minimum width=1cm, text width=4cm, minimum height=1cm,text centered, draw=black]
\tikzstyle{block2} = [rectangle, minimum width=1cm, text width=3cm, minimum height=1cm,text centered, draw=black]
\tikzstyle{arrow} = [thick,->,>=stealth]
\tikzstyle{arrow2} = [thick,|]

\tikzstyle{startstop} = [rectangle, rounded corners, minimum width=3cm, minimum height=1cm,text centered, draw=black, fill=red!30]
\tikzstyle{process} = [rectangle,rounded corners, minimum width=3cm, minimum height=1cm, text centered, draw=black, fill=orange!30]
\tikzstyle{arrow} = [thick,->,>=stealth]

\usepackage{microtype}

\usepackage{soul}

\aclfinalcopy 

\title{BreakingBERT@IITK at SemEval-2021 \\Task 9 : Statement Verification and Evidence Finding with Tables}

\author{
    Aditya Jindal$^{*}$ \qquad   
    Ankur Gupta$^{*}$ \qquad
    Jaya Srivastava$^{*}$ \\
  \large{\textbf{Preeti Menghwani}}$^{*}$ \qquad   
  \large{\textbf{Vijit Malik}}$^{*}$ \qquad
  \large{\textbf{Vishesh Kaushik}}\thanks{\quad Authors equally contributed  to this work. Names in alphabetical order.}\\
  \large{\textbf{Ashutosh Modi}} \\
{Indian Institute of Technology Kanpur (IIT Kanpur)} \\
{\tt \{adityaji, ankugupt, jayasri\}@iitk.ac.in}\\
{\tt \{priti, vijitvm, kvishesh\}@iitk.ac.in}  \\
{\tt ashutoshm@cse.iitk.ac.in}  \\
}

\date{}

\begin{document}
\maketitle
\begin{abstract}
Recently, there has been an interest in factual verification and prediction over structured data like tables and graphs. To circumvent any false news incident, it is necessary to not only model and predict over structured data efficiently but also to explain those predictions. In this paper, as part of the SemEval-2021 Task 9,  we tackle the problem of fact verification and evidence finding over tabular data. There are two subtasks. Given a table and a statement/fact, subtask A determines whether the statement is inferred from the tabular data, and subtask B determines which cells in the table provide evidence for the former subtask. We make a comparison of the baselines and state-of-the-art approaches over the given SemTabFact dataset. We also propose a novel approach CellBERT to solve evidence finding as a form of the Natural Language Inference task. We obtain a 3-way F1 score of 0.69 on subtask A and an F1 score of 0.65 on subtask B. 

\end{abstract}
\section{Introduction}
Textual Inference, also known as natural language inference \cite{bowman2015large}, plays an important role in the study of natural language understanding and semantic representation. Due to the unprecedented amount of information generated over the internet, it becomes essential for machines to comprehend new information based on previous knowledge. Recent social events like political elections and pandemic spread have also shown the need for intelligent fact-checking systems that majorly depends on textual Inference over the scientific data. 

Though Textual Inference is well explored, the current works mainly deal with unstructured Evidence in the form of sentences \cite{inproceedings}. Verification under structured and semi-structured Evidence, such as tables, graphs, and databases, remains unexplored. Tables are ubiquitous in documents and presentations for concisely conveying important information; however, Inference on structured data like tables or graphs is much more difficult than simple text format due to complex structure and non-universal schema for the representation of data. Though recently, there has been work on Tabular Inference problems ( \citealp{zhong2020logicalfactchecker}; \citealp{cho2018adversarial}; \citealp{sun2018knowledgeaware}; \citealp{2019TabFactA}; \citealp{eisenschlos2020understanding}; \citealp{pasupat-liang-2015-compositional}; \citealp{wang-etal-2018-neural-question}; \citealp{neeraja-etal-2021-infotabskg}; \citealp{gupta-etal-2020-infotabs};) explaining the prediction, evidence finding is still an unexplored area. 

Through the SemEval-2021 Task 9 \cite{wang2021semeval} we have tried to solve the Tabular Inference problem over scientific tables by providing an answer as well as a solution to our reasoning. In other words, given the structured table data and statement, we aim to classify the statement as entailed, unknown (neutral), or contradiction. In addition, we also aim to classify each cell of the table whether it is relevant or irrelevant in making the aforementioned prediction.  Our contribution is three-fold:
\begin{itemize}
    \item We perform an empirical study of current state-of-the-art models on the SemTabFact dataset for the task of statement verification (see Section \ref{sec:resultssubA}).
    \item We implement TableSciBERT, TableRoBERTa and develop a heuristic-based classifier. We achieve a 3-way F1 score of 0.69 on statement verification by ensembling TableSciBERT and TAPAS with our heuristic method (see Section \ref{subsec:subtaskA}). 
    \item We propose a new model CellBERT, for the task of Evidence finding from the tables. We achieve an F1 score of 0.65 on Evidence finding by ensembling CellBERT with our heuristic-based approach (See Section \ref{subsec:subbapproach}).
\end{itemize}


\noindent The code for all our experiments and pre-trained models are available on GitHub\footnote{\href{https://github.com/vijit-m/TablEval}{https://github.com/vijit-m/TablEval}}.

\section{Background}
\subsection{Related Work}
\label{sec:relatedwork}

 Recently, \citet{2019TabFactA} proposed TabFact, a dataset with 16k Wikipedia tables and 118k human-annotated natural language statements, labeled as either ENTAILED or REFUTED. The authors proposed the TableBERT model for the task of fact-checking. TableBERT uses the pre-trained BERT \cite{devlin2018bert} model and fine-tunes it using the TabFact dataset as a simple NLI task by linearizing the table along with the fact. The linearized table is then concatenated with the statement which after tokenization is given as input to the BERT model which is used for binary classification to predict the nature of the statement.
 
 The paper also proposed LPA (Latent Program Algorithm) to formulate the table fact-checking as a program synthesis problem. LPA uses reinforcement learning to optimize the task reward of this structured prediction problem directly, as was done in Neural Symbolic Machines (NSM) \cite{liang2016neural}.
\citet{zhong2020logicalfactchecker} used the combination of the linguistic and symbolic reasoning integrated with an understanding of a given table's structural format.

\begin{figure}[]
    \includegraphics[width=0.44\textwidth]{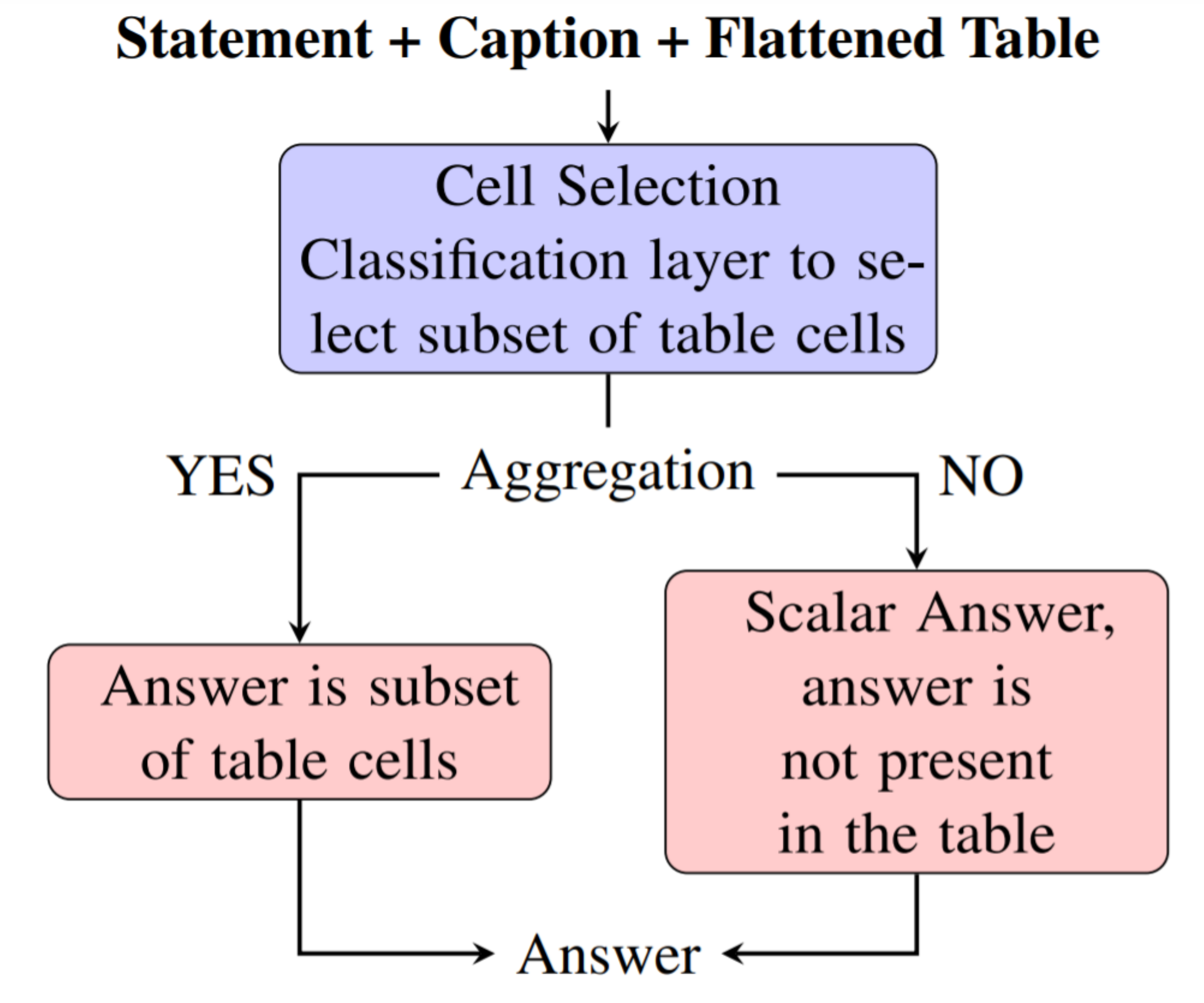}
    \caption{ Flow chart explaining the functioning of TAPAS model using the two classification layers.}
    \label{fig:tapaschart}
\end{figure}

\citet{herzig-etal-2020-tapas} developed the TAPAS model that performs question-answering over tables without generating logical forms. It uses weak supervision and predicts the answer by selecting table cells and optimally applying aggregation operators (for example: count, sum, average etc.) to the selected cells . An input instance to the model is the combination of the tokenized question and flattened table, separated by an \texttt{[SEP]} token (see Figure \ref{fig:tapaschart}). In addition to BERT embeddings, TAPAS incorporates the table's structural information via row, column, and rank embeddings. Since TAPAS uses flattened tables, it also suffers from the limitation of self-attention computation over long input sequences like BERT. Due to this reason, it fails to capture information over large tables or /databases containing/ multiple tables. \par
Besides this, the model's expressivity is limited to a form of aggregation over a few cells of the table; hence, it fails to handle questions requiring multiple aggregation operations properly.

\noindent Recently, \citet{eisenschlos2020understanding} adapted TAPAS for the task of fact-checking. They introduced two intermediate pre-training tasks learned from the MASK-LM model. The first task is based on counterfactual statements, generated by creating one positive and one negative from every relevant Statement extracted from Wikipedia statements and tables. The second task is based on synthetic statements that generate a sentence by sampling from a set of logical expressions. 

Let $S$ and $T$ represent the statement/fact and the table, respectively, which are given as input to the model. Furthermore, let $E_{S}$ and $E_{T}$ represent the corresponding input embeddings. The sequence of statement and the table given by $E = [E\textsubscript{\texttt{[CLS]}}$; $E\textsubscript{S}; E\textsubscript{\texttt{[SEP]}}; E\textsubscript{T}]$ is passed through the transformer, $f$ and a contextual representation is obtained for every token. The entailment probability $P(S|T)$ is modeled using a single hidden layer neural network obtained by computing the output of $\texttt{[CLS]}$ token:

\begin{equation}
    P(S|T) = MLP(f_{\textit{[CLS]}}(E))
\end{equation}


To handle large size tables, Table pruning is done using Heuristic exact match (HEM). In this method, the columns are ranked by a relevance score and added in order of decreasing relevance. Columns that exceed the maximum input length are skipped. 
\label{sec:length}

Our task for statement verification differs from the works mentioned above in the aspect that we have a third label, `unknown', where the table fails to provide any information to infer the Statement. Moreover, no prior work has been performed concerning the task of evidence finding.

\subsection{Problem Defintion} 

The problem statement is articulated around the following two related subtasks.

\noindent\textbf{Subtask A - Table Statement Support:}
Given a statement/fact, some of which will be directly adapted from the linking text, and a table, determine whether the table's information supports the Statement. In this classification problem, a statement is assigned one of the following labels:
\begin{enumerate}
\item[$\bullet$] \textbf{Fully Supported: } Statement is supported by data found within the table (denoted by 1).
\item[$\bullet$] \textbf{Refuted: } Statement is contradicted by table (denoted by 0).
\item[$\bullet$] \textbf{Unknown: } Not enough information in table to assess statement veracity (denoted by 2).
\end{enumerate}

\noindent Mathematically, the problem can be described as, given a table $\mathbb{T}$ and a statement $\mathbb{S}$, we need to learn a mapping $\mathbb{F}_{A}$ to the output $y_{A}$, where $y$ $\in \{0,1,2\}$. See examples in \cref{table:kysymys} and \cref{table:2}.







\noindent\textbf{Subtask B - Relevant Cell Selection:}
Given a statement and a table, determine which table cells form relevant evidence for the Statement (if any). A table cell is evidence for a statement if it helps support or refute a part of the statement. In this subtask each cell of the table is assigned the following labels:



\begin{enumerate}
\item[$\bullet$] \textbf{Relevant: } the cell must be included (denoted by 1).
\item[$\bullet$] \textbf{Irrelevant:} the cell must not be included (denoted by 0).
\end{enumerate}


\noindent Mathematically, each cell $x_{ij} \in \mathbb{T}$ (where $i$, $j$ correspond to row and column number respectively), needs to be assigned a value $y_{B}$ $\in$ $\{0,1\}$. See examples in \cref{table:kysymys} and \cref{table:2}.


\begin{table}[h!]
\begin{center}
\begin{tabular}{|p{2.8cm}|p{2cm}|p{1.5cm}|} 
\hline 
\cellcolor{blue!15}\textbf{Body Sensation}& \cellcolor{red!15}\textbf{Agoraphobic} & \cellcolor{red!15}\textbf{Pleasant}\\ 
\hline 
\cellcolor{red!15}number & \cellcolor{red!15}museum  & \cellcolor{red!15}lovely  \\ 
\hline 
\cellcolor{blue!15}palpitation & \cellcolor{red!15}shop  & \cellcolor{red!15}happiness \\ 
\hline
\cellcolor{red!15}heartbeat & \cellcolor{red!15}boat  & \cellcolor{red!15}Joyous   \\ \hline 
\end{tabular} 
\caption{\footnote{based on example 0001.html with light edits} A sample table and statement with correct results for subtask B. \textcolor{blue!65}{violet}: Relevant Cell,  \textcolor{red!65}{red}: Irrelevant Cell}
\label{table:kysymys}
\end{center}
\end{table}
\vspace{0.2cm}


\begin{table}[H]
\begin{adjustbox}{width=0.9\columnwidth,center}
\begin{tabular}{|c|c|}
\hline
\textbf{Statement}                 & \textbf{Label} \\ \hline
Palpitation is a bodily sensation  & Supported      \\ \hline
Joyous and boat have same strength & Unknown        \\ \hline
Lovely is an agoraphobic situation & Refuted        \\ \hline
\end{tabular}
\end{adjustbox}
\caption{Statements and Labels corresponding to Sub-task A}
\label{table:2}

\end{table}
\label{sec:length}

\noindent\textbf{Corpus Collection:} The training and testing data is sourced from
open-access scientific articles with tables using
APIs provided by Science Direct for data mining.
The data is procured in XML format and each table is also provided in image format since the size and styling of
table contents are useful in understanding the table structure. Two separate datasets (with varying complexity) are provided, one in which the statements are automatically-generated, the second one where statements are generated manually by humans. The automatically generated statements are relatively more complex \cite{wang2021semeval} and more in number as compared to the Manually generated statements. 

\noindent\textbf{Annotation Process: }Each statement in the SemTabFact dataset is adapted from existing text and verified by at least one human reader. Multiple readers verify a smaller
proportion to
assess inter-annotator agreement. 

\noindent\textbf{Data Preprocessing:} The Tables in the dataset have multiple sub-columns, unlike the Wikipedia tables in the TabFact dataset.
The Training dataset had few errors like label classification errors and Grammatical errors. To overcome this, data is preprocessed and cleaned before feeding it into the models. During the Data Cleaning, statements with no labels are removed, and subcolumns are interpolated to handle the tables' complex header structure.\par
The subcolumns are also merged with the table headers in the case of multiple Table headers that improve the results. The tables are provided with surrounding text along with the captions. During preprocessing, the surrounding text is combined with captions which together serve as captions to the subtask A model. Refer to Section \ref{subsec:subtaskA} for our system description for subtask A. (See Appendix \ref{supp:preprocessing} for preprocessing details). We also discuss the result of this step on our model's performance in Section \ref{res:EA:subA}


\vspace{1mm}
\begin{table*}[]
\centering
\begin{tabular}{|c|c|c|c|c|c|c|}
\hline
\textbf{Source} & \textbf{\#Tables} & \textbf{\#Entailed} & \textbf{\#Refuted} & \textbf{\#Unknown} & \textbf{\#Relevant} & \textbf{\#Irrelevant} \\ \hline
Manual          & 981               & 2818                & 1688               & 0                  & 0                   & 0                     \\ \hline
Auto-gen.  & 1980              & 92136               & 87209              & 0                  & 1039058             & 15467957              \\ \hline
Development     & 52                & 250                 & 213                & 93                 & 3048                & 28495                 \\ \hline
Test            & 52                & 274                 & 248                & 131                & 3458                & 26724                 \\ \hline
\end{tabular}
\caption{Dataset statistics for different datasets within SemTabFacts.}
\label{tab:my-table}
\end{table*}


\section{System Desciption}

\subsection{Subtask A}
\label{subsec:subtaskA}
\begin{figure*}[]
    \centering
    \includegraphics[width=0.9\textwidth]{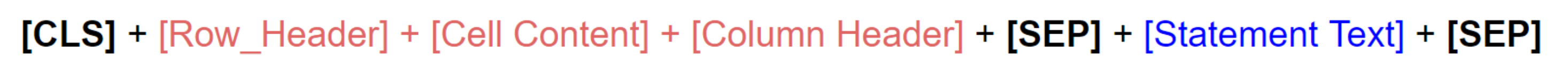}
    \caption{Approach for Subtask B}
    \label{fig:subbappraoch}
\end{figure*}

\label{sec:length}

Our proposed model for subtask A is an ensemble of TableSciBERT, TAPAS, and a heuristic-based approach. We first use a similarity-based approach to predict and segregate `unknown' statements, and then we predict the remaining statements using an ensemble of TableSciBERT and TAPAS. Given a statement, $S$ and table, $T$, in order to classify whether $S$ is `unknown' or not, we first calculate the similarity score between $S$ and each cell $C \subset T$ using equation \ref{eqn:simcell}. Here, $Sim$ is a similarity function, C is the content of the cell, $s_{i}$ is i-th token of the statement S after removing stop words and $c_{j}$ is the j-th token of cell (for handling multi-word cells). We use the nltk library \cite{bird2009natural} to tokenize the statement and cell contents. The similarity function takes as input two tokens and outputs the similarity between them in 0 to 1 (higher score representing more similar). For each token in the statement, we first iterate through all the tokens of the cell and compute the maximum of scores obtained by $Sim$ function for each token in the cell and the particular token of the statement. We then sum it over all the tokens of the statement to compute the score, $s_{c}$. We obtain the aggregated score, $s_{s}$ over the whole table $T$ by adding score $s_{c}$, of each cell $c \subset T$ (see equation \ref{eqn:simstat}). Refer to Section \ref{sec:subaexperiments} for more information about the types of similarity functions we experimented with.

\begin{align}
        s_{c} &= \sum_{i}^{}\maxB_{j}(Sim(c_{j},s_{i})), s_{i} \subset S, c_{j} \subset C
        \label{eqn:simcell}\\
        s_{s} &= \sum_{c}s_{c}, \textrm{  where  } C \subset T \label{eqn:simstat}
\end{align}

We use $s_{s}$ as the similarity score between statement $S$ and table $T$. If $s_{s} < \lambda_{a}$, we label the statement as `unknown' where $\lambda_{a}$ is a hyperparameter. If $s_{s} \geq \lambda_{a}$ we proceed with the two-way classification using our ensemble of TableSciBERT and TAPAS.

Note that these TableSciBERT and TAPAS models were fine-tuned upon two labels only (viz-a-vis Entailed and Refuted). The ensemble is done by applying weighted average upon prediction probabilities of TAPAS and TableSciBERT. TableSciBERT (from TableBERT) was developed by replacing the BERT base model with SciBERT \cite{beltagy2019scibert}. SciBERT is a pre-trained language model based on BERT, which is fine-tuned upon large scale scientific data. Since the SemTabFact dataset is from scientific articles, using SciBERT makes sense intuitively. 


TAPAS and TableSciBERT were trained on the training set (both autogenerated and manual dataset) and the development set. The table pruning method using the Heuristic exact match (HEM) was applied for the large complex tables in an Autogenerated dataset to handle the input embedding size. We also experimented with other transformers models like BioBERT \cite{lee2020biobert}, CovidBERT\footnote{\href{https://huggingface.co/gsarti/covidbert-nli}{https://huggingface.co/gsarti/covidbert-nli}} but SciBERT gave the best result. Furthermore, we experimented with training them as a 3-label classifier as well. Refer to the Section \ref{sec:Experiments} for details.

For training TableSciBERT and other table transformer 3-label variants we augmented the training data with statements having ‘unknown’ label, in order to balance the scarcity of unknown labels in the provided data. The augmentation for a table was done by randomly sampling statements from other tables provided in the dataset having ‘entailed’ or ‘refuted’ as the true label.  The motivation behind using this strategy was that these statements served as statements with an ‘unknown’ label for this table. During sampling, we ensured that entailed and refuted statements are equal in number to prevent any bias. 


Overall, given a table $\mathbb{T}$ and statement $\mathbb{S}$, the complete pipeline is a two-step process:
\begin{itemize}
    \item We first perform the binary classification of whether $\mathbb{S}$ is `unknown' or not using the Similarity heuristic. If the predicted label is `unknown', it is taken as the final prediction; otherwise, we proceed to the next step. 
    \item We use the ensemble of TAPAS and TableSciBERT models to predict the `entailed' or `refuted' label of $\mathbb{S}$. Here, $\mathbb{S}$ are the statements that were NOT classified as `unknown' in the previous step.
\end{itemize}

\subsection{Subtask B}
\label{subsec:subbapproach}
For subtask B, we developed an ensemble of two different techniques: 
\begin{itemize}
\item \textbf{CellBERT}: We propose a new method CellBERT as a BERT-base model that is fine-tuned upon an individual cell-based Natural Language Inference Task. We preprocess the training data given to us in Subtask B (which consists of only auto-generated statements) to generate NLI input samples of the form described in Figure \ref{fig:subbappraoch}. Mathematically, given a statement $S$ and a table $T$, we need to label each cell $c$ in the table as relevant or irrelevant. For CellBERT, each cell's label is individually determined along with the supporting information of row and column headers. In other words, if the coordinates of a cell $c$ are given by $(x,y)$, where $x$ is the row number and $y$ is the column number, the coordinates of the row header and column header cells are given by $(x,1)$ and $(1,y)$ respectively.\par
The motivation behind using row and column headers information is that these capture the `type' of data present in the cell. The combination of the row header, the cell, and the column header's contents represent the NLI task's premise. The hypothesis is taken as the statement provided. Note that using this approach, we ended up with around a million data points to train upon. Due to the unavailability of adequate computational resources, we restricted to using only 0.1\% of the preprocessed training data.
    
    \item \textbf{Similarity}: We observed that Scientific Tables contained many entities for which pre-trained word embeddings are unavailable, and thus supervised approaches like CellBERT, fails to capture the required relationship. To overcome this, we used a cell-wise similarity algorithm, which calculates the score $s_{c}$ of each cell $C$ with the statement $S$ same as in equation \ref{eqn:simcell}. 
    We used $s_{c}$ as the similarity score between statement $S$ and cell $C$. If $s_{c} < \lambda_{b}$ we label the cell as `irrelevant', where $\lambda_{b}$ is a hyperparameter. Otherwise, we label the cell as `relevant'.
    

\end{itemize}

\label{sec:Experiments}
\section{Experiments}
\subsection{Subtask A}
\label{sec:subaexperiments}
Following the TabFact's TableBERT, we fine-tuned our own TableBERT model on the SemTabFact dataset for Subtask A. We also experimented with mutations of TableBERT by using RoBERTa (TableRoBERTa), XLNet (TableXLNet), and SciBERT (TableSciBERT) as well. We also experimented with implementing BiGRU layers on top of these table transformers. All our experiments were conducted using PyTorch \cite{NEURIPS2019_9015} Deep Learning library.

We experimented with a dataset ($D_{1}$) which contained the Manual dataset along with the Autogenerated statements. A caveat to $D_{1}$ was that autogenerated statements that have common tables with the manual dataset were only used. This was done because every model we trained upon only on the manual dataset was overfitting. The overfitting was due to an insufficient number of statements, i.e., $4056$ in the Manual Dataset. After preparing the dataset $D_{1}$, we had a total of 72k statements.\\
For the similarity-based approach, we manually experimented with various non-semantic similarity approaches like edit distance and binary-matching\footnote{We define binary matching score between two tokens $t_{1}$ and $t_{2}$ as 1 if the lower-cased, lemmatized and stemmed form of both the tokens is the same otherwise it is taken as 0.} as well as embedding space-based semantic similarity approaches by first computing the word vector. We computed word vectors using GloVe \cite{pennington-etal-2014-glove}, BERT \cite{devlin2018bert} and RoBERTa \cite{liu2019roberta} . Cosine similarity and euclidean distance was used to compute similarity between two vectors. The non-semantic based binary-matching approach outperformed others upon the validation set; therefore, we used it to evaluate our results on the test set.
In our final submitted model, SciBERT was fine-tuned for $3$ epochs upon the combined dataset $D_{1}$ and development dataset, with learning rates as $5e-5$, $5e-6$ and $1e-6$ for each epoch. Batch size was kept as $6$ with maximum sequence length of $512$ tokens. TAPAS was fine tuned upon the auto-generated, manual and development datasets separately for $6$, $12$ and $5$ epochs respectively. Learning rate was kept the same as $2e-5$ for each epoch with maximum sequence length as $512$. Dropout probability was set to $0.07$. For ensemble, we used weights 0.7 and 0.3 for TAPAS and TableSciBERT respectively. We set the hyperparameter $\lambda_{a}$ as 2 in our similarity heuristic.

\subsection{Subtask B}
\label{sec:subbexperiments}

For Subtask B, we preprocessed the dataset into input samples as shown in Figure \ref{fig:subbappraoch}, and fine-tuned a BERT base model. Since the number of statements given corresponding to the auto-generated dataset is large and also accounting for the fact that each cell in the table is a separate input example, the number of tuples of (cell, statement) were very large (over ten million data points). Therefore, only 0.1\% of all tuples ($\approx$ 30k data points) were used to train CellBERT, and the rest of the data was discarded. Note that the 0.1\% of the data that we selected to train CellBERT was kept completely balanced with respect to true labels. We also experimented with including and not-including header information during fine-tuning as well. See table \ref{Table:res_subb}.


Here as well, for the similarity-based approach, we manually experimented with the same non-semantic similarity approaches like edit distance and binary-matching as well as embedding space-based semantic similarity approaches we used in Section \ref{sec:subaexperiments}. In Subtask B too, the non-semantic based binary-matching approach outperformed others upon the validation set, hence we used the same to evaluate our results on the test set. For our final model, the hyperparameter $\lambda_{b}$ was set to 1. For CellBERT we fine-tuned a BERT base model for 5 epochs with batch size 16 and learning rate as $2e-5$.

\section{Results}
\subsection{Subtask A}
\label{sec:resultssubA}
The organizers use two evaluation metrics for subtask A:
\begin{itemize}
\item \textbf{3-way-F1:} This is a standard precision/recall evaluation (Three-Way) of a multi-class classification that evaluates whether each statement is classified as Entailed / Refuted / Unknown.
\item \textbf{2-way-F1:} The second evaluation method is a Two Way method in which statements with the unknown ground truth label are not taken into consideration.
\end{itemize}
In both methods, first F1 scores are calculated for each table, which is then averaged across all tables for the final F1 score. The results of subtask A on the test set are shown in \cref{tab:res_suba}:\\
\begin{table}[H]
\centering
\resizebox{7.7cm}{!}{
\begin{tabular}{|c|c|c|}
\hline
\textbf{Model} &
  \textbf{\begin{tabular}[c]{@{}c@{}}Task A 2-way\\ F1\end{tabular}} &
  \textbf{\begin{tabular}[c]{@{}c@{}}Task A 3-way \\ F1\end{tabular}} \\ \hline
\textbf{\begin{tabular}[c]{@{}c@{}}(TAPAS+TableSciBERT)-\\ 2-Label+Similarity\end{tabular}} &
  \textbf{0.7681} &
  \textbf{0.6931} \\ \hline
BERT-3-Label                                                            & 0.5963 & 0.5295 \\ \hline
\begin{tabular}[c]{@{}c@{}}TAPAS -2-Label\\ + Similarity\end{tabular}   & 0.7547 & 0.6824 \\ \hline
\begin{tabular}[c]{@{}c@{}}SciBERT-2-Label \\ + Similarity\end{tabular} & 0.6172 & 0.5534 \\ \hline
RoBERTa-3-Label                                                         & 0.6186 & 0.5271 \\ \hline
\begin{tabular}[c]{@{}c@{}}(RoBERTa+BiGRU)\\ 3-Label\end{tabular}       & 0.5986 & 0.5113 \\ \hline
\end{tabular}
}
\caption{Test Data Result (Average F1-scores) for subtask A}
\label{tab:res_suba}
\end{table}

\begin{figure}[H]
\centering
\includegraphics[width=0.5\textwidth]{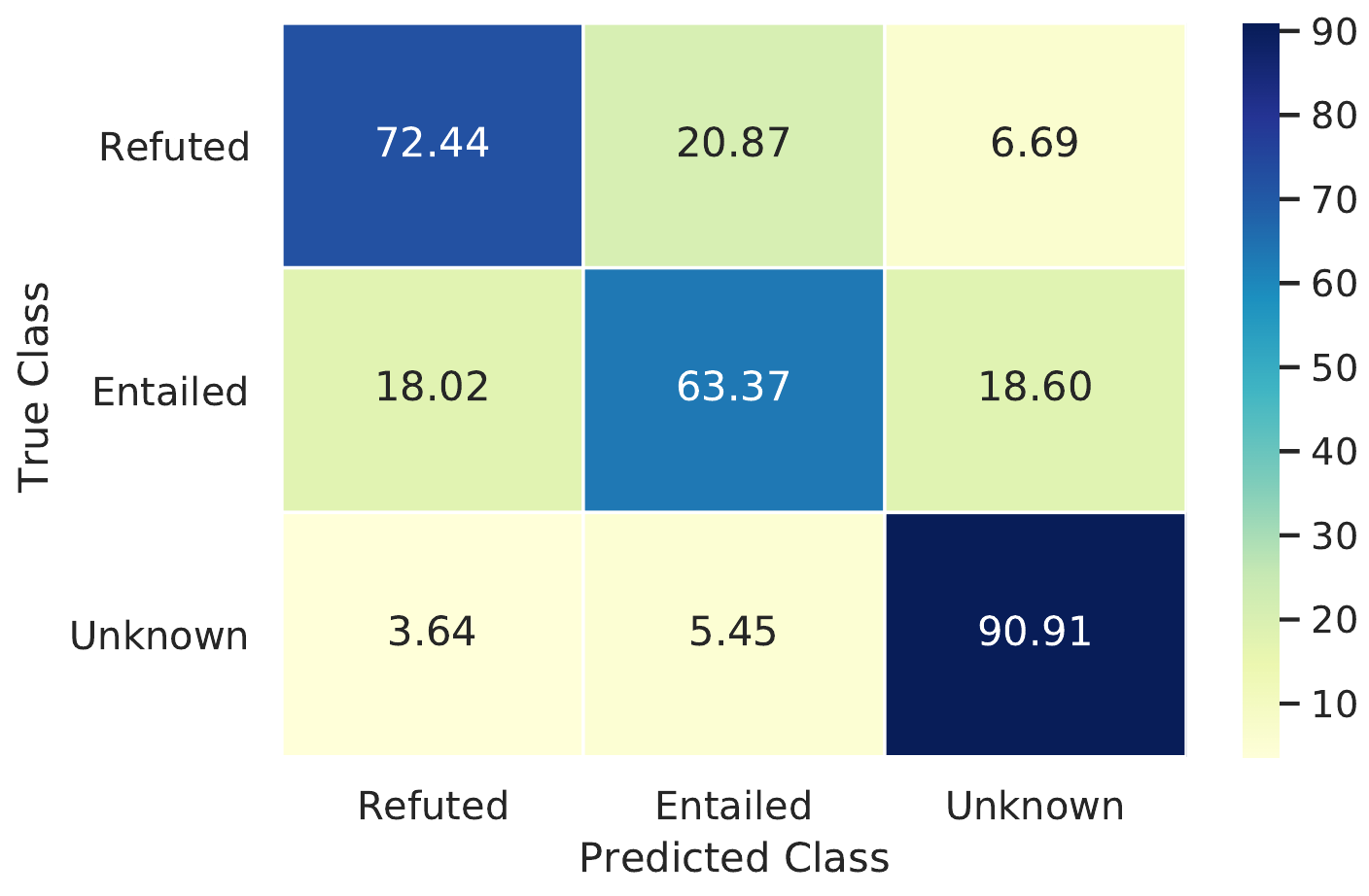}
\caption{Confusion Matrix of Testset on
(Tapas+TableSciBERT)-2-Label+Similarity.}
\label{fig:histostatstats}
\end{figure}

Confusion Matrix of Test set on (TAPAS+TableSciBERT)-2-Label+Similarity is shown in \cref{fig:histostatstats}
The best results for subtask A were obtained using (TAPAS+TableSciBERT)-2-Label+ Heuristic-based Similarity with 0.768 as 2-way F1 and 0.693 as 3-way-F1. TAPAS-2-Label+ Heuristic Based Similarity also gave second-best results with 0.755 as 2-way F1 and 0.682 as 3-way-F1. We have experimented with other table transformers like RoBERTa, BERT, and XLNet, but none of these gave promising results. Models like RoBERTa-3-Label, BERT-3-Label have been trained on the three labels and gave 0.527 and 0.529, respectively, as 3-way-F1. From the experiments, we observed that the models trained on 3 label classification performed poorly in classifying statements with `unknown' label.

\subsection{Subtask B}
The metric used by the organizers calculates the recall and precision for each cell, with ``relevant" cells as the positive category. Similar to Task A, the score is averaged over all statements in each table first, before calculating average across all tables. The results for subtask B on test data are shown in \cref{Table:res_subb}. We obtained the best F1 of 0.6517. We have also shown the results of other variants of Cell-BERT, which are classified on the basis of row and column headers (see Table \ref{Table:res_subb}).

\begin{table}[H]
\centering
\resizebox{7.5cm}{4.5cm}{
\begin{tabular}{|c|c|c|}
\hline
\textbf{S.No} & \textbf{Method} & \textbf{F1} \\ \hline
1. & Similarity & 0.6414 \\ \hline
2. & CellBERT & 0.5380 \\ \hline
3. & CellBERT - DevT & 0.6465 \\ \hline
\textbf{4.} & \textbf{\begin{tabular}[c]{@{}c@{}}CellBERT - DevT\\ + Similarity Ensemble\end{tabular}} & \textbf{0.6517} \\ \hline
5. & \begin{tabular}[c]{@{}c@{}}CellBERT \\ (only cell context)\end{tabular} & 0.4891 \\ \hline
6. & \begin{tabular}[c]{@{}c@{}}CellBERT \\ (cell+row header \\ information)\end{tabular} & 0.5213 \\ \hline
7. & \begin{tabular}[c]{@{}c@{}}CellBERT \\ (cell+column header \\ information)\end{tabular} & 0.5199 \\ \hline
8. & \begin{tabular}[c]{@{}c@{}}CellBERT \\ (cell+row+column\\ header information)\end{tabular} & 0.5380 \\ \hline
\end{tabular}
}
\caption{Test Data Result (Average F1-scores) for subtask B}
\label{Table:res_subb}
\end{table}


\section{Error Analysis}
\label{res:EA:subA}

\noindent\textbf{Subtask A: }
On analyzing the training dataset, we realized that many statements require aggregation methods like sum, count, max, and min over the tabular data to determine whether a statement is entailed or refuted (or if it is `unknown'). It requires a symbolical understanding of the text that can not be understood using simple NLI based approaches like Table-BERT and other table transformers. On the other hand, TAPAS outperforms other models primarily due to pre-training on the corpus of synthetic and counterfactual statements, as discussed in Section \ref{sec:relatedwork}.


We noticed that TableSciBERT performed well as compared to other Table Transformers on subtask A. It makes intuitive sense as the dataset is created from scientific texts and consequently has scientific statements. The organizers also mentioned that the training and testing data is sourced from open access scientific articles with tables using APIs provided by Science Direct for data mining in the task description.\par
Further, we improved the F1 score on the TAPAS model by 2\% by using the multiple header merging technique (See Appendix \ref{supp:preprocessing}). The reason being that the merged headers have more semantic information as they contain the sub-headers too. We have explained the pre-processing step of tables with multiple headers and sub-columns in the \nameref*{my_appendix}.  For 3-way classification, the table transformers gave unsatisfactory results, for example, RoBERTa 3-Label with F1 of 0.527 primarily due to two reasons, first we need complete domain knowledge of the related table to tag a statement to be unknown, and second, no data were available for the unknown label in the training set. We obtain 3-way F1 of 0.693, with (TAPAS+SciBERT)-2-Label+Similarity model, as our similarity heuristic was successfully able to classify 91\% of the statements predicted unknown as true unknown labels. This might be possible because many unknown statements in the development and test set are completely independent of the table given.\\



\noindent\textbf{Subtask B: } We noticed that CellBERT performs best when header information is included while fine-tuning. An interesting point to note is that CellBERT trained only on autogenerated dataset provides an F1 score of 0.538, whereas when it was later fine-tuned upon the Development Dataset which had Manual based dataset (CellBERT-DevT), it boosted the results on the test set to F1 score of 0.646. The reason being the Test Dataset being a Manual based dataset, whereas the earlier CellBERT model was trained on the more complex autogenerated dataset.\par
The heuristic-based Similarity approach was performing surprisingly well on the test set as well, giving us a comparable result of 0.641 F1. Although both the approaches are entirely different, the scores are comparable. This motivated us to analyze each method's predictions, but we could not come up with a convincing hypothesis for what might be the reason for this observation. However, we noticed a common trend in both methods. When the number of cells in a table is large, and the number of relevant cells is very less, both models failed to identify relevant cells in such cases. In other words, both models had difficulty in identifying true positives.

\section{Conclusion}
This paper attempts a solution to an under-explored but essential problem: Statement Verification and Evidence Finding with Tables. There have been various works related to a binary classification of statements. Still, evidence finding for these classifications is a difficult and novel challenge. 
We are successfully able to present an ensemble of TAPAS model, table transformer-based TableSciBERT, and similarity heuristic trained for subtask A with 2-way F1 of 0.768 and 3-way F1 of 0.693. In subtask B, we implemented the CellBERT - DevT+ Similarity Ensemble method as our best model with an F1 score of 0.652.

In the future, we plan to progress in implementing new models that can tackle both linguistic and symbolic reasoning. We aim to extend the TAPAS model to 3 labels, requiring large data for training in unknown labels for good results.
In the case of subtask B, we are planning to experiment with other NLI techniques and models. Besides, we will be looking into using more data for training CellBERT and other NLI models as well.



\bibliographystyle{acl_natbib}
\bibliography{anthology,acl2021}

\clearpage
\newpage

\appendix


\section*{Appendix} \label{my_appendix}

\section{Preprocessing of the Multiple Header files} \label{supp:preprocessing}

\mbox{}\vspace{3px}
\newline
The provided dataset had many Tables with multiple headers and subcolumns, as shown in \cref{tab:app1}.

Since most of our models take a single header as input only and with an equal number of columns in every row, we had to convert such tables to suit our input type. There were two processes involved for preprocessing such Tables.

\textbf{Intrapolation}: For one table, we would calculate the maximum numbers of columns in the Table and then interpolate all other rows with less columns to eventually have an equal number of columns in every row. Resulting \cref{tab:app2} obtained by interpolation of \cref{tab:app1} is shown below

\vspace{-0.3cm}
\begin{table}[H]
\begin{tabular}{|c|c|c|c|c|}
\hline
\multirow{2}{*}{ExperMatter} & \multicolumn{2}{c|}{UserB} & \multicolumn{2}{c|}{UserC} \\ \cline{2-5} 
                             & Base1        & Base2       & Base1        & Base2       \\ \hline
Gold                         & 5.6          &             & 7            & 8           \\ \hline
Silver                       & 3.4          & 6.7         & 8.0          & 6.7         \\ \hline
\end{tabular}
\caption{Table with multiple headers and subcolumns }
\label{tab:app1}
\end{table}
\vspace{-0.5cm}

\begin{table}[H]
\begin{tabular}{|p{1.8cm}|p{1cm}|p{1cm}|p{1cm}|p{1cm}|}
 \hline
ExperMatter &UserB&UserB&UserC&UserC\\
 \hline
 & Base1&Base2&Base1&Base2\\
 \hline
  Gold& 5.6& & 7&8 \\
  \hline
  Silver&3.4&6.7&8.0&6.7   \\
 \hline
\end{tabular}
\caption{Table after Intrapolation}
\label{tab:app2}
\end{table}
\vspace{-0.5cm}
\textbf{Header Merging}: Since the input our model has to be a single header file we had to merge such rows as shown below. Final table obtained after Preprocessing is shown in \cref{tab:app3}.

\begin{table}[H]
\begin{tabular}{|p{1.8cm}|p{1cm}|p{1cm}|p{1cm}|p{1cm}|}
 \hline
ExperMatter &UserB Base1&UserB Base2&UserC Base1&UserC Base2 \\
 \hline
  Gold& 5.6& & 7&8 \\
  \hline
  Silver&3.4&6.7&8.0&6.7   \\
 \hline
\end{tabular}
\caption{Table after preprocessing}
\label{tab:app3}
\end{table}

\begin{table*}[h]
\begin{tabular}{|l|c|c|c|c|c|c|}
\hline
\multirow{2}{*}{\textbf{Model}} & \multirow{2}{*}{\textbf{\begin{tabular}[c]{@{}c@{}}Train\\ set\end{tabular}}} & \multirow{2}{*}{\textbf{\begin{tabular}[c]{@{}c@{}}Dev \\ set\end{tabular}}} & \multicolumn{4}{c|}{\textbf{Metrics (On Dev set)}}                                         \\ \cline{4-7} 
                                &                                                                               &                                                                               & \textbf{Precision} & \textbf{Recall} & \textbf{F1}   & \textbf{Accuracy (\%)} \\ \hline

\textbf{TAPAS}              & \textbf{Auto}                                                              & \textbf{Auto}                                                               & \textbf{99.4}      & \textbf{94.7}   & \textbf{97.1}  & \textbf{94.32}        \\ \hline
\textbf{TAPAS}                    & \textbf{Auto+Manual}    & \textbf{Manual}                                                                            & \textbf{82.6}               & \textbf{71.5}        & \textbf{76.7}        &\textbf{70.83}               \\ \hline
TableBERT            & Auto                                                             & Auto                                                               & 87.5    & 85.9   & 86.7 & 86.00         \\ \hline
TableRoBERTa                    & Auto                                                                          & Auto                                                                          & 63.5               & 63.1            & 63.3          & 64.05                   \\ \hline
TableRoBERTa + BiGRU            & Auto                                                                          & Auto                                                                          & 64.7               & 63.4            & 66.0          & 67.13                   \\ \hline
TableSciBERT                    & Auto                                                                          & Auto                                                                          & 71.0               & 69.8            & 70.4          & 70.74                  \\ \hline

TableBERT                       & Auto                                                                          & Manual                                                                        & 58.8               & 58.2            & 58.5          & 58.95                   \\ \hline
TableRoBERTa                    & Auto                                                                          & Manual                                                                        & 52.9               & 50.7            & 51.8          & 51.95                   \\ \hline
\end{tabular}
\caption{Results on the Development set created by us}
\label{tab:ourressubA}
\end{table*}
\newpage

\section{Other Experiments with training data} \label{supp:other_exps}
We have also provided the results for 2-way F1 on the development set, which we have created out of the training dataset in \cref{tab:ourressubA} were trained on only two labels.
We divided both the Manual and Auto-generated training data into Train and Dev set with a split of $90:10$, respectively. Various combinations of Train and Dev data were used to train different models.
The models given in \cref{tab:ourressubA} were trained on only two labels as both of the training datasets do not contain unknown labels, and this is also the reason why we have not shown 3-way F1.\\
TAPAS model gave us the best here too, with an F1 score of 97.1 when the Autogenerated train set and Dev set were used, while it gave us the best F1 score of 76.7 when the Autogenerated + Manual train set and Manual Dev set were used.  We have not included any of the models trained only on the Manual train dataset because the provided Manual data is very small in numbers and the results of any model trained on manual train set were not very promising. Clearly TAPAS outperforms all the models and gives the best results on both the Datasets.


\end{document}